\documentclass{article} % For LaTeX2e
\usepackage[utf8]{inputenc}
\usepackage{amsmath, amssymb, latexsym}
\usepackage{sidecap}
\usepackage{algorithm} 
\usepackage{algpseudocode} 

\usepackage{tikz}
\usetikzlibrary{decorations.pathreplacing}
\usepackage[english]{babel}
\newtheorem{theorem}{Theorem}
\usepackage[utf8]{inputenc}
\usepackage{mathtools}
\usepackage{latex_template,times}
\usepackage{hyperref}
\usepackage{url}
\usepackage{graphicx}
\usepackage{csvsimple}
\usepackage{multirow}
\usepackage{adjustbox}
\usepackage{subcaption}
\usepackage{pgfplots}
\pgfplotsset{compat=1.14}
\usepackage{setspace}
\usepackage{amsmath} 

\usepackage{authblk}
\author[1]{Pengwei Yang}
\author[1]{Chongyangzi Teng}
\author[2]{Jack George Mangos}
\affil[1]{School of Computer Science, The University of Sydney, Sydney NSW 2000, Australia}
\affil[2]{School of Medicine, University of New South Wales, Kensington NSW 2052, Australia}
\date{}                     %% if you don't need date to appear
\title{Establishment of Neural Networks Robust to Label Noise}

\nipsfinalcopy % Uncomment for camera-ready version

\begin{document}

% \author{Pengwei Yang\and Angel Teng \and
% Jack Mangos}
% The University of Sydney

% Anonymous Author(s) \\
% Affiliation \\
% Address \\
% \texttt{email} \\

\maketitle

\begin{abstract}
Label noise is a significant obstacle in deep learning model training. It can have a considerable impact on the performance of image classification models, particularly deep neural networks, which are especially susceptible because they have a strong propensity to memorise noisy labels. In this paper, we have examined the fundamental concept underlying related label noise approaches. A transition matrix estimator has been created, and its effectiveness against the actual transition matrix has been demonstrated. In addition, we examined the label noise robustness of two convolutional neural network classifiers with LeNet and AlexNet designs. The two FashionMINIST datasets have revealed the robustness of both models. We are not efficiently able to demonstrate the influence of the transition matrix noise correction on robustness enhancements due to our inability to correctly tune the complex convolutional neural network model due to time and computing resource constraints. There is a need for additional effort to fine-tune the neural network model and explore the precision of the estimated transition model in future research.
\end{abstract}

\section{Introduction}
\label{gen_inst}

Deep learning has led to significant breakthroughs in computer vision and image processing and has been applied to a variety of applications, including handwriting recognition \cite{dutta_improving_2018}, satellite image classification \cite{lemaire_fpga-based_2020}, IoT crowdsourcing \cite{yang2023monitoring}, and medical picture classification for illness detection \cite{zhang_medical_2019}. In the majority of applications, the existence of label noise in the training datasets is one of the primary obstacles to training these deep learning models.

Label noise is the presence of labels for classification data in which the labels do not precisely reflect the instance's content. Because generating large labelled datasets can be resource heavy and costly, dataset labelling is frequently performed by non-professional labellers or automated systems with minimal or no expert supervision \cite{guo_ms-celeb-1m_2016} \cite{zhong_unequal-training_2019}. Additionally, in some subject categories, constructing labelled datasets is intrinsically challenging due to the lack of assurance in the image content. For instance, collections of medical photographs frequently exhibit significant observer variability and, consequently, substantial instance-dependent label noise \cite{ju_improving_2022}, as images pose a genuine diagnostic difficulty for specialists.

In general, there are three types of label noise: class-independent (sometimes known as "uniform") noise, class-dependent noise, and class and instance-dependent noise \cite{frenay_classification_2014}. This research examines class-dependent label noise, in which labels are flipped randomly with a probability dependent on the class. This class noise rate will be referred to as the "flip rate" in the study.

Label noise can dramatically impair classification model performance \cite{fang_learning_2019}. Deep neural networks can be more susceptible to label noise because they have a larger propensity to memorise noisy labels, which can have a negative impact on their capacity to generalize \cite{zhang_understanding_2017} \cite{algan2020label}. Due to the difficulty of noisy labels, learning from datasets with noisy labels has been the focus of extensive research in recent years.

In Section 2 of this research, we  first summarise and compare a variety of label noise techniques described in the literature. In section 3, we will describe the structure of the neural networks we will employ in our own experimental work, and in Section 4, we will provide our own experimental strategy for addressing label noise. Using backwards noise correction and a transition matrix, we analyse the label noise robustness of two distinct neural networks (LeNet and AlexNet) and compare their performance using this method. The experiment is conducted using three datasets; the first two are grayscale photos from FashionMINIST, each with distinct flip rate and a transition matrix. The third dataset consists of images with unknown flip rates from CIFAR. We independently estimate the transition matrix for the second dataset. There are three label classes in all three datasets. In Section 5, we give a conclusion of the whole research.

\section{Related work}
\label{headings}

There are many methods available in the literature that have been used to address the problem of label noise. This section serves as an overview and an analysis of their relative strengths and weaknesses. These methods described in the literature can be roughly divided into three separate groupings, as described by Karimi et al \cite{karimi_deep_2020}.

The first grouping is methods focusing on model selection or design – designing models that are more robust to label noise, through selecting the model, loss function, or training procedures. For example, in building simple classifiers, naïve Bayes and random forest methods are known to be more robust to label noise than decision trees and support vector machines \cite{nettleton_study_2010}; similarly, bagging is a superior method to boosting in noise robustness \cite{dietterich_experimental_2000}. These methods have the advantage of not requiring modifications of the training data in order to succeed, and therefore are potentially more generalizable between instances of model use than other methods. However, they restrict the types of models available for data processing, which therefore decreases the highest potential accuracy.

Set of studies that also propose adding a “noise layer” to the end of deep learning models, as an additional buttress against the effect of noisy labels. Multiplication of the output of a deep learning model with a transition matrix (as explored in this paper) falls under this category. However, other methods also exist – for example, adding a noise layer to train an adversarial network under label noise \cite{segata_scalable_2009}. This method has the benefit of ensuring the learned classifier from the noisy dataset is guaranteed to be risk consistent with the clean dataset \cite{brodley_identifying_1996}. However, the downside is that it has to use the noise transition matrix, which is not always available in real-world applications.

The loss function is also a way to improve robustness to label noise. Many studies keep all other components static (model design, data, etc.) and change only the loss function to determine the effect. In regard to deep learning, it has been shown that the mean absolute value of error (MAE), is tolerant to label noise, whereas cross-entropy loss and mean square error are not \cite{wilson_reduction_2000}; qualitatively this is because MAE treats all examples more equally than do the other examples, which may place more emphasis on discordant noisy instances. There have consequently been studies on improving both the MAE and the cross-entropy loss functions to improve the performance of both on noisy data \cite{zhang_methods_2009} \cite{veit_learning_2017}.  Similarly, the 0-1 label loss function is known to be more robust than smoother functions (such as log-loss or squared loss) \cite{karimi_deep_2020}.

Second grouping is methods that aim to reduce the label noise within the training data itself. There is an enormous spectrum of these methods. Some rely on the presence of clean data, such as a small clean test dataset, in order to identify where the noisy labels lie to facilitate their removal. One popular method in this subcategory is to train a classifier using the noisy data and then remove mislabelled data samples based on the classifier predictions \cite{northcutt_learning_2017}. A similar method was explored by Veit et al., where two CNNs were trained in parallel – one trained on a small clean dataset to remove noisy data, another on a larger noisy dataset (cleaned) to perform the main classification task; this method was shown to be superior to training on the noisy dataset followed by fine-tuning on the clean dataset \cite{thekumparampil_robustness_2018}.

Methods of removing noisy labels without a clean dataset for comparison also exist. K-nearest neighbours and ensemble voting on mislabelled samples between differently trained/ or structured classifiers have been used to coarsely eliminate samples deemed noisy, however it has been identified that these methods are prone to removing too many instances \cite{ghosh_robust_2017} \cite{wang_imae_2019}. There are also more computationally intensive methods available, such as using a leave-one-out framework to identify the overall impact on classification of a single instance as an indication of whether it is mislabelled or not, and excluding those that sit above a certain threshold \cite{thulasidasan_combating_2019}. Similarly, the rank pruning method attempts to identify data points with labels that are deemed “confident”, based on the assumption that data samples for which the predicted probability is close to one are more likely to have correct labels, and uses these alone to update the classifier \cite{jiang_mentornet_2017}. The universal issue with these methods is that they may introduce bias into the classifier by inappropriately removing samples misidentified as noise, as genuine outliers, or otherwise complex instances, may be mislabelled and removed inappropriately.

Third grouping is based on adapted training procedures specifically for the presence of label noise. For example, curriculum learning is based on training a model with initially easy examples, and escalating the difficulty, such that the model has a formidable baseline from which to encounter noisy data; this method has been shown to significantly improve image classification on CIFAR-100 when a “mentor” network trains a “student” network \cite{malach_decoupling_2017}. One other proposed method involved training two separate networks, both initialised randomly, and only updating the parameters when the predictions of the two models differed, as a way to “screen” for noisy labels \cite{xia_are_2019}. The benefit of this strategy is that it is resistant to overfitting incorrect labels, and, particularly toward the end of training, will prefer harder examples with correct labels. The downside of these training procedure methods, however, is they are more theoretically and computationally involved, and can introduce significant amounts of additional complexity.

\section{Methodology}
To build our classifiers, we chose two deep neural network models which were computationally feasible within the scope of this paper. Consequently, we chose to use the LeNet-5 and AlexNet as our baselines. As for data preprocessing, apart from resizing input data to fit different neural network architectures, we also used random flip as a data augmentation technique, with a flip probability of 0.5. In addition, we utilised the cross-entropy loss function with an SVD optimizer through this paper (we respectively set the momentum to 0.9 and the weight decay parameter to 0.00005). In terms of the neural network model tuning process, we began by setting the batch size to 128, the learning rate to 0.1, and the epoch number to 15. Then, according to the training accuracy and validation accuracy, we first tuned the learning rate (from 0.1 to 0.0001) and then the batch size to encourage convergence to the global minimum. However, some models did not have the opportunity to converge to the optimum due to the high computation complexity.

\subsection{Lenet-5}
\begin{figure}[htb]
% \begin{subfigure}{1\linewidth}\centering
\begin{center}
% \advance\leftskip-3cm
\includegraphics[width=0.6\textwidth]{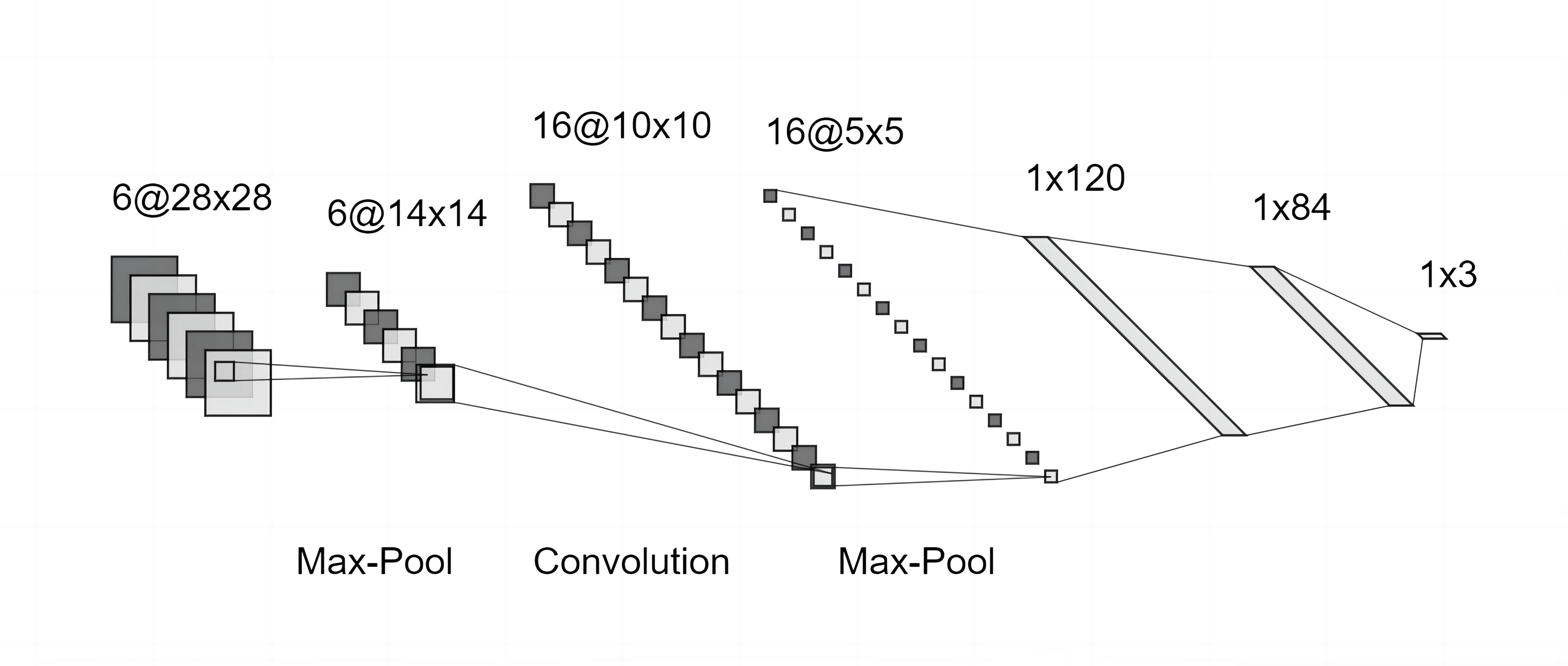}
\caption{Graphical representation of the LeNet-5 architecture}
{\label{lenet}}
\end{center}
% \end{subfigure}
\end{figure}

LeNet-5 is a classical neural network architecture. It contains five layers - two composite convolutional layers, and three fully connected layers. Each composite layer contains three components: a convolutional layer, a maximum pooling layer, and a nonlinear activation function. The network uses three layers of multilayer perceptrons as the combined classification layer. Input data is required to be of shape 32x32, and as such our team used the transform library from PyTorch to resize the input data. The output of the model was set from ten to three, to match the number of classes in all of the input datasets. Fig.~\ref{lenet} demonstrates this structure graphically. 

\subsection{AlexNet}
\begin{figure}[htb]
% \begin{subfigure}{1\linewidth}\centering
\begin{center}
% \advance\leftskip-3cm
\includegraphics[width=0.6\textwidth]{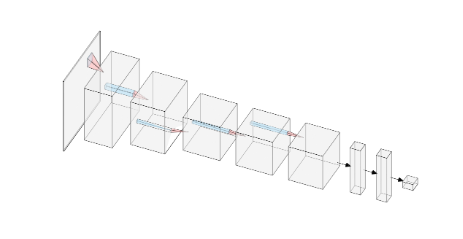}
\caption{Our AlexNet model, demonstrating input and eight sequential convolutional layers.}
{\label{alex}}
\end{center}
% \end{subfigure}
\end{figure}

The second neural network used by our team is AlexNet. AlexNet is another convolutional neural network which is eight layers deep. It was originally designed to perform classification on the ImageNet database; as such, it is capable of outputting up to 1000 output classes, and its input data size is 227x227. Again, to facilitate the use of this model, we used the PyTorch transform library to reshape our data to fit the required input. Fig.\ref{alex} demonstrates the structure of our AlexNet model. 

Due to its larger size, AlexNet contains many more weight and bias parameters to train than LeNet-5. In order to avoid the model becoming computationally infeasible, AlexNet includes dropout of neurons at each convolutional layer with a set probability, to reduce computation time while including minimal bias. While the standard AlexNet uses local response normalisation, our group used batch normalisation to reduce the effects of exploding and vanishing gradients. 

\subsection{Label noise learning using backward learning algorithm}
The focus of this paper is on instance-independent and class-dependent label noise (CCN). In order to perform learning in the setting of noisy label data, we chose to adopt the backward learning (also called the backward correction procedure) algorithm \cite{patrini_making_2017}. The basic idea of backward learning is to output the dot product of the inverted transition matrix with the output of the neural network (a set of conditional probabilities). The corrected loss of this output then equals the output of the clean data under expected label noise \cite{patrini_making_2017}. The detailed backward learning algorithms can be seen in Figure 3. 

As explained above, this method requires a transition matrix, which is defined as a matrix of probabilities that, as dictated by the class-specific probabilities, clean labels flip into noisy labels. This transition matrix is, in normal scientific work, estimated; for this paper, two of three required transition matrices were provided.  

\begin{theorem}
Suppose that the transition matrix T is non-singular. Given a loss $\ell $, backward corrected loss is defined as:
$$\ell^{\leftarrow}(\hat{p}(y|x))=T^{-1}\ell(\hat{p}(y|x))$$
Then, the loss correction is unbiased, i.e. :
$$\forall x,\: \mathbb{E}_{\tilde{y}|x}\ell^{\leftarrow}(y,\hat{p}(y|x))=\mathbb{E}_{\tilde{y}|x}\ell(y,\hat{p}(y|x))$$
and therefore the minimizers are the same:
$$\underset{\hat{p}(y|x)}{argmin} \mathbb{E}_{x,\tilde{y}}\ell^{\leftarrow}(y,\hat{p}(y|x))=\underset{\hat{p}(y|x)}{argmin} \mathbb{E}_{x,y}\ell(y,\hat{p}(y|x))$$
\end{theorem}

By using the backward learning method, the noisy posterior (derived from the neural network output) can be transformed into an estimate of the clean posterior. Theoretically, after building a neural network based on CCN noisy labels and using backward learning, the classifier may perform better when using clean test datasets.

A visualisation of the backward learning algorithm is provided in Figure~\ref{Backward}. In this visualisation, T represents the transition matrix, x from 1 to N is the input instances and y from 1 to C represents conditional probabilities of C classes. The transformation of the noisy posterior to the clean posterior is shown to be the result of the product of the noisy posterior with the inverse transition matrix T. 

\begin{SCfigure}[12]
	\centering
    \begin{tikzpicture}[shorten >=1pt]
        \tikzstyle{unit}=[draw,shape=circle,minimum size=1.15cm]
 
        \node[unit](x0) at (0,3.5){$x_1$};
        \node[unit](x1) at (0,2){$x_2$};
        \node(dots) at (0,1){\vdots};
        \node[unit](xn) at (0,0){$x_N$};
 
        \node[unit](y1) at (4,2.5){$y_1$};
        \node(dots) at (4,1.5){\vdots};
        \node[unit](yc) at (4,0.5){$y_C$};
 
        \draw[->] (x0) -- (y1);
        \draw[->] (x0) -- (yc);
 
        \draw[->] (x1) -- (y1);
        \draw[->] (x1) -- (yc);
 
        \draw[->] (xn) -- (y1);
        \draw[->] (xn) -- (yc);
        \draw   (5,1.5) node[xshift=100]{$y = P(\tilde{Y}|X) \rightarrow y_b = P(Y|X) = T^{-1}y$};
 
        \draw [decorate,decoration={brace,amplitude=10pt},xshift=-4pt,yshift=0pt] (-0.5,4) -- (0.75,4) node [black,midway,yshift=+0.6cm]{input layer};
        \draw [decorate,decoration={brace,amplitude=10pt},xshift=-4pt,yshift=0pt] (3.5,3) -- (4.75,3) node [black,midway,yshift=+0.6cm]{output layer};
    \end{tikzpicture}
\caption[Figure]{}
\label{Backward}
\end{SCfigure}
The necessary limitation of this backward approach is that it requires a transition matrix, and, as such, its performance also relies on the quality of the transition matrix. In the real world setting, this is almost never available. In addition, although our group chose to use the backward learning approach (as it was an achievable algorithm within the scope of the paper) past research has compared the performance between backward approach to forward correction approach and has found that the latter often performs better \cite{patrini_making_2017}. From our anecdotal survey of the literature, our team has also noticed that the forward correction approach is more commonly used than the backward approach.
\subsection{Estimating the transition matrix with anchor points}
When the true transition matrix is not available, such as for the CIFAR dataset in this paper, the transition matrix needs to be estimated. This estimation can be done by estimating the noisy class probabilities via anchor points. Anchor points are instances belonging to a specific class with probability as close to one as possible \cite{patrini_making_2017}. 

The approach using anchor points is illustrated in the formula below. Given an instance ($ x^0 $) that we can confidently classify into a specific class (class 0), the use of this class probability as essentially = 1 means we can calculate a portion of the transition matrix. The corollary is therefore that if there exist anchor points for each class, the full transition matrix can be estimated. As for CIFAR, which is modified to 3 classes, to calculate the transition matrix, it is assumed that there exists three anchor points. This assumption is then fulfilled by taking those instances with the highest probability as surrogate anchor points. 
{
    \begin{align}
        \begin{bmatrix}
        P(\tilde{Y}=0|X=x^0) \\
        P(\tilde{Y}=1|X=x^0) \\
        P(\tilde{Y}=2|X=x^0) 
        \end{bmatrix} &=
        \begin{bmatrix}
        P(\tilde{Y}=0|Y=0) &  P(\tilde{Y}=0|Y=1) & P(\tilde{Y}=0|Y=2)\\
        P(\tilde{Y}=1|Y=0) &  P(\tilde{Y}=1|Y=1) & P(\tilde{Y}=1|Y=2)\\
        P(\tilde{Y}=2|Y=0) &  P(\tilde{Y}=2|Y=1) & P(\tilde{Y}=2|Y=2)
        \end{bmatrix} 
        \begin{bmatrix}
        P(Y=0|X=x^0) \\
        P(Y=1|X=x^0) \\
        P(Y=2|X=x^0)
        \end{bmatrix} \\
        &=\begin{bmatrix}
        P(\tilde{Y}=0|Y=0) &  P(\tilde{Y}=0|Y=1) & P(\tilde{Y}=0|Y=2)\\
        P(\tilde{Y}=1|Y=0) &  P(\tilde{Y}=1|Y=1) & P(\tilde{Y}=1|Y=2)\\
        P(\tilde{Y}=2|Y=0) &  P(\tilde{Y}=2|Y=1) & P(\tilde{Y}=2|Y=2)
        \end{bmatrix} 
        \begin{bmatrix}
        1 \\
        0 \\
        0
        \end{bmatrix} \\
        &=\begin{bmatrix}
        P(\tilde{Y}=0|Y=0) \\
        P(\tilde{Y}=1|Y=0) \\
        P(\tilde{Y}=2|Y=0) 
        \end{bmatrix}    
    \end{align}
}

We defined a function called “estimator” to implement the aforementioned process. The purpose of this function was to find instances with the highest conditional probabilities that could be classified into one class. This then allowed our group to find three satisfactory anchor points for the three classes in the dataset, and therefore build our transition matrix. The aforementioned algorithm can be seen in Algorithm 1. 

\begin{algorithm}
	\caption{Transition Matrix Estimator}
	\hspace*{\algorithmicindent} \textbf{Input$P(\tilde{Y}|X)$} \\
    \hspace*{\algorithmicindent} \textbf{Output$\:T_{00},T_{10},T_{20},T_{10},T_{11},T_{12},T_{20},T_{21},T_{22}$}
	\begin{algorithmic}[1]
		\For {$P(\tilde{Y}|x)$ from $P(\tilde{Y}|X)$}
        \If {$\exists \: P(\tilde{Y}|x^0)>T_{00}$}
    	\State$T_{00}\leftarrow P(\tilde{Y}=0|X=x^0)$
    	\State$T_{10}\leftarrow P(\tilde{Y}=1|X=x^0)$
    	\State$T_{20}\leftarrow P(\tilde{Y}=2|X=x^0)$
    	\EndIf
    	\If {$\exists \: P(\tilde{Y}|x^1)> T_{11}$}
    	\State$T_{01}\leftarrow P(\tilde{Y}=0|X=x^1)$
    	\State$T_{11}\leftarrow P(\tilde{Y}=1|X=x^1)$
    	\State$T_{21}\leftarrow P(\tilde{Y}=2|X=x^1)$
    	\EndIf
    	\If {$\exists \: P(\tilde{Y}|x^2)> T_{22}$}
    	\State$T_{02}\leftarrow P(\tilde{Y}=0|X=x^2)$
    	\State$T_{12}\leftarrow P(\tilde{Y}=1|X=x^2)$
    	\State$T_{22}\leftarrow P(\tilde{Y}=2|X=x^2)$
     	\EndIf
		\EndFor
	\end{algorithmic} 
\end{algorithm}

\section{Experiment}
\subsection{Evaluation metrics}

The performance of each classier will be evaluated with the top-1 accuracy metric,
that is,
$$top-1 \, _{accuracy}=\frac{\text{number of correctly classified examples}}{\text{total number of test examples}}*100 \%$$

\subsection{Experiment setup}
\begin{figure}[htb]
\begin{subfigure}{1\linewidth}\centering
\begin{center}
\includegraphics[width=1\textwidth]{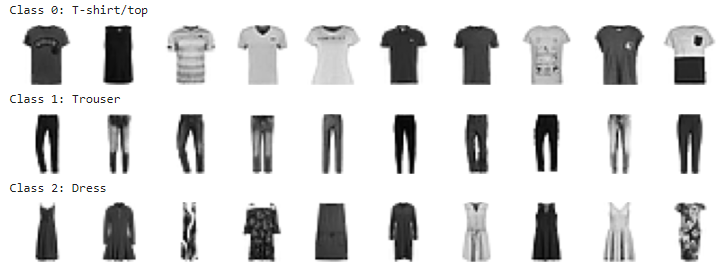}
\caption{FashionMINIST images with clean label}
\end{center}
\end{subfigure}
\begin{subfigure}{1\linewidth}\centering
\begin{center}
\includegraphics[width=1\textwidth]{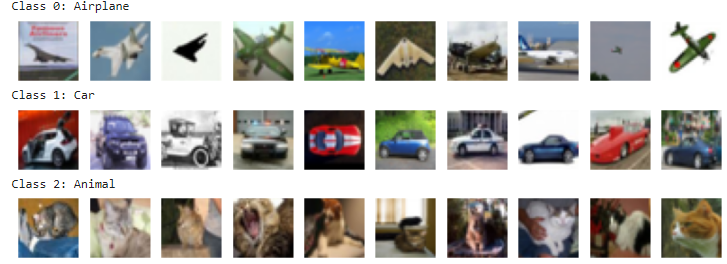}
\end{center}
\caption{CIFAR images with clean label}
\end{subfigure}
\caption{\label{fig:example_image}Image examples from MNIST and CIFAR}
\end{figure}

In the experiments, we used three datasets with noisy labels, FashionMINIST0.5, FashionMINIST0.6 and CIFAR. The first two dataset are created from the FashionMINIST which contains fashion items and its category. The third dataset is created from the commonly used CIFAR dataset. Figure~\ref{fig:example_image} gives example images of the datasets and table~\ref{table_summary} provides a summary of the datasets.

We designed two baseline classification models, convolutional neural networks classifiers using LeNet and convolutional neural networks classifiers using AlexNet.

The noise transition matrix is provided for the first two datasets, listed in section\ref{Transition Matrix}\textit{Transition Matrix} below. The transition matrix for CIFAR is unknown, as such a transition matrix estimator is built for CIFAR.

We then applied backward noise correction using noise transition matrix on the two classification models for all three dataset.

Two classification models with and without transition matrix were assessed using top-1 accuracy metric for their robustness.

For rigorous performance evaluation, we have trained each classifier 5 times with different training and validation sets generated by random sampling. The average and standard deviation of the test accuracy is papered. 

%examples of images

\begin{table}[h]
\caption{Summary of the datasets}
\label{table_summary}
\begin{center}
% \begin{adjustbox}{width=1\textwidth}
\begin{tabular}{|c||c|c|c|c|}
\hline
 Dataset &  Training &  Test &   Class & Image size\\
\hline\hline
FashionMINIST05 & 18000 & 3000 & 3 & $28 \times \:28$\\
FashionMINIST06 & 18000 & 3000 & 3 & $28 \times \:28$\\
CIFAR & 15000 & 3000 & 3 & $32 \times \:32 \times 3$\\
\hline

\end{tabular}
% \end{adjustbox}
\end{center}
\end{table}

\subsubsection{Provided Transition Matrix}
\label{Transition Matrix}
\;\;\;\;\;\;\;\;
FashionMINIST0.5 =
$ \begin{bmatrix}
0.5 & 0.2 & 0.3\\
0.3 & 0.5 & 0.3\\
0.3 & 0.3 & 0.5\\
\end{bmatrix}$ \;\;\;\;\;\;\;\;\;\;\;\;
FashionMINIST0.6 =
$ \begin{bmatrix}
0.4 & 0.3 & 0.3\\
0.3 & 0.4 & 0.3\\
0.3 & 0.3 & 0.4\\
\end{bmatrix}$

\subsection{Experiment results and discussion}

\subsubsection{Estimated Transition Matrix}

To validate the effectiveness of our transition matrix estimator, we estimated the transition estimation method using both LeNet and AlexNet and compared them against the true transition matrices of the first two datasets. 

The estimator highly depends on the posteriors from neural networks, when neural network changes the transition matrix would be changed accordingly. That will induce instability for our label noise learning. The estimated transition matrix shown below is produced by one iteration run to demonstrate the results. Results are rounded to 2 decimal places for easy reading.

\subsubsubsection{\textbf{FashionMINIST0.5}}
\\

\:\:\:\:LeNet =
$ \begin{bmatrix}
0.55 & 0.24 & 0.33\\
0.25 & 0.5 & 0.16\\
0.21 & 0.27 & 0.51\\
\end{bmatrix}$, \:\:Error$_{LeNet}$ = 
$ \begin{bmatrix}
-0.05 & -0.04 & -0.03\\
0.05 & 0.00 & 0.04\\
-0.01 & 0.03 & 0.09\\
\end{bmatrix}$

AlexNet = 
$ \begin{bmatrix}
0.56 & 0.17 & 0.30\\
0.27 & 0.55 & 0.18\\
0.17 & 0.28 & 0.52\\
\end{bmatrix}$, Error$_{AlexNet}$ = 
$ \begin{bmatrix}
-0.06 & 0.03 & 0.00\\
0.03 & -0.05 & 0.02\\
0.03 & 0.02 & 0.08\\
\end{bmatrix}$
\subsubsubsection{\textbf{FashionMINIST0.6}}

\:\:\:\:LeNet =
$ \begin{bmatrix}
0.42 & 0.26 & 0.28\\
0.28 & 0.44 & 0.28\\
0.30 & 0.30 & 0.44\\
\end{bmatrix}$, \:\:Error$_{LeNet}$ = 
$ \begin{bmatrix}
-0.02 & 0.04 & 0.02\\
0.02 & -0.04 & 0.02\\
0.00 & 0.00 & -0.04\\
\end{bmatrix}$

AlexNet =
$ \begin{bmatrix}
0.40 & 0.29 & 0.30\\
0.31 & 0.45 & 0.32\\
0.30 & 0.27 & 0.38\\
\end{bmatrix}$, Error$_{AlexNet}$ = 
$ \begin{bmatrix}
0.00 & 0.01 & 0.00\\
-0.01 & -0.05 & -0.02\\
0.00 & 0.03 & 0.02\\
\end{bmatrix}$
 
\subsubsubsection{\textbf{CIFAR}}
\\
Estimated Transition Matrix for CIFAR using\:\:\:\: LeNet = 
$ \begin{bmatrix}
0.46 & 0.30 & 0.35\\
0.28 & 0.40 & 0.25\\
0.26 & 0.30 & 0.40\\
\end{bmatrix}  $

Estimated Transition Matrix for CIFAR using AlexNet = 
$ \begin{bmatrix}
0.49 & 0.30 & 0.32\\
0.27 & 0.40 & 0.27\\
0.24 & 0.31 & 0.41\\
\end{bmatrix}  $

The estimated transition matrix using both LeNet and AlexNet matched the provided true transition matrix for both FashionMINIST0.5 and FashionMINIST0.6. LeNet has slightly better accuracy demonstrated by smaller variances. However, without repeating the estimation several rounds, it is hard to make concrete decisions between the models due to the volatility of the results.

Nevertheless, the small variances between the estimated and true matrices has demonstrated our transition matrix estimator is effective thus can be applied to the CIFAR dataset.

\begin{table}[h]
\caption{Average test accuracy \% with standard deviation}
\label{iteration demonstration}
\begin{center}
% \begin{adjustbox}{width=1\textwidth}
\begin{tabular}{|c||c|c|c|}
\hline
Model & FashionMINIST0.5 & FashionMINIST0.6 & CIFAR\\
\hline\hline
LeNet  & 91.96($\pm$1.14)& 75.63($\pm$4.13) & 49.21($\pm$4.61)\\
LeNet-Backward &\textbf{92.07($\pm$0.94)} & \textbf{86.53($\pm$1.22)} & 52.18($\pm$7.85)\\
AlexNet & 90.23($\pm$0.81) & 82.28($\pm$3.07) & 51.44($\pm$8.21)\\
AlexNet-Backward &90.44($\pm$0.53) & 85.94($\pm$2.77) & \textbf{59.73($\pm$4.05)}\\

% LeNet  & \textbf{92.7(0.59)} & 86.31(1.59) & 49.49(9.53)\\
% LeNet-Forward & 91.53(1.48) & 77.23(4.1) & 49.63(1.2)\\
% AlexNet & 90.29(0.71) & \textbf{87.12(1.96)} & \textbf{57.72(3.43)}\\
% AlexNet with transition matrix & 90.52(0.65) & 70.42(11.09) & 53.08(2.44)\\

\hline
\end{tabular}
% \end{adjustbox}
\end{center}
\end{table}

\begin{figure}[htb]
\begin{minipage}[b]{0.30\textwidth}
\begin{center}
\begin{tikzpicture}
    \begin{axis}
    [
   width=0.7\textwidth,
   scale only axis,
   ymin=30, ymax=100,
   ylabel={Top 1 Accuracy (\%)},
   xlabel={Iteration},
   axis lines*=left,
   legend style ={ at={(3,2),
        anchor=north west, draw=black, 
        fill=white,align=left,
        nodes={scale=0.3, transform shape}}
    }]
\addplot[orange,dashed] table [x=Iteration, y=LeNet, col sep=comma] {iteration_results_1.csv};
\addlegendentry{LeNet}
\addplot[orange,thick,mark=square*] table [x=Iteration, y=LeNet_T, col sep=comma] {iteration_results_1.csv};
\addlegendentry{LeNet - Backward}
\addplot[blue,dashed] table [x=Iteration, y=AlexNet, col sep=comma] {iteration_results_1.csv};
\addlegendentry{AlexNet}
\addplot [blue,thick,mark=square*]  table [x=Iteration, y=AlexNet_T, col sep=comma] {iteration_results_1.csv};
\addlegendentry{AlexNet - Backward}
\end{axis}
\end{tikzpicture}
\caption{\label{fig:Iteration_FashionMINIST0.5}FashionMINIST0.5}
\end{center}
\end{minipage}
\hfill
\begin{minipage}[b]{0.3\textwidth}
\begin{center}
\begin{tikzpicture}
    \begin{axis}
    [
   width=0.7\textwidth,
   scale only axis,
   ymin=30, ymax=100,
   ylabel={Top 1 Accuracy (\%)},
   xlabel={Iteration},
   axis lines*=left,
   legend style ={}]
\addplot[orange,dashed] table [x=Iteration, y=LeNet, col sep=comma] {iteration_results_2.csv};
\addplot[orange,thick,mark=square*] table [x=Iteration, y=LeNet_T, col sep=comma] {iteration_results_2.csv};
\addplot[blue,dashed] table [x=Iteration, y=AlexNet, col sep=comma] {iteration_results_2.csv};
\addplot [blue,thick,mark=square*]  table [x=Iteration, y=AlexNet_T, col sep=comma] {iteration_results_2.csv};
\end{axis}
\end{tikzpicture}
\caption{\label{fig:Iteration_FashionMINIST0.6}FashionMINIST0.6}
\end{center}
\end{minipage}
\hfill
\begin{minipage}[b]{0.3\textwidth}
\begin{center}
\begin{tikzpicture}
    \begin{axis}
    [
   width=0.7\textwidth,
   scale only axis,
   ymin=30, ymax=100,
   ylabel={Top 1 Accuracy (\%)},
   xlabel={Iteration},
   axis lines*=left,
   legend style ={}]
\addplot[orange,dashed] table [x=Iteration, y=LeNet, col sep=comma] {iteration_results_3.csv};
% \addlegendentry{LeNet}
\addplot[orange,thick,mark=square*] table [x=Iteration, y=LeNet_T, col sep=comma] {iteration_results_3.csv};
% \addlegendentry{LeNet with Transition Matrix }
\addplot[blue,dashed] table [x=Iteration, y=AlexNet, col sep=comma] {iteration_results_3.csv};
% \addlegendentry{AlexNet}
\addplot [blue,thick,mark=square*]  table [x=Iteration, y=AlexNet_T, col sep=comma] {iteration_results_3.csv};
% \addlegendentry{AlexNet with Transition Matrix}
\end{axis}
\end{tikzpicture}
\caption{\label{fig:Iteration_CIFAR}CIFAR}
\end{center}
\end{minipage}
\end{figure}

During the experiments, after injecting the backward learning algorithm into our neural networks, we found that the learning rate needed to be smaller compared with the original neural networks without backward learning to ensure high accuracy.

All four models have demonstrated their robustness for dataset FashionMINIST0.5. LeNet has slightly better performance than AlexNet. When the transition matrix is used, LeNet has relatively more volatility compared to AlexNet but no strong difference in performance was seen. 

Compared to FashionMINIST0.5, the performance deteriorated slightly for FashionMINIST0.6 for all models. This is as expected due to the noisy label rate of FashionMINIST0.6 being higher. Therefore, it is unsurprising that there was an observed decrease in model performance.

Performance for CIFAR for all four models is lower compared the two FashionMINIST datasets. There are many potential causes of this. It could be due to a higher noise rate, due to error within the estimated transition matrix, or due to the fact that the classification problem was more complex because the images were coloured. 

Although theoretically using a transition matrix could improve robustness to label noise, we have not observed a positive effect for both LeNet and AlexNet on the two FashionMINIST0.6 and CIFAR datasets. Some positive impact was observed for AlexNet for FashionMINIST0.5 using the backward algorithm, however the improvement is small and, given the size, it is hard to make any concrete conclusions. In addition, we have noticed, using the backward algorithm does increase volatility of model performance demonstrated by large standard deviations. 

Contradictory results observed in our experiment could be due to the following four reasons, 1) our neural model is not finely tuned, 2) The calculation of inversing the transition matrix might have introduced additional error into the results, 3) Adding a transition matrix could further increase model complexity and impact performance if not tuned properly, and 4) The transition matrix is not accurately estimated for CIFAR dataset which could impact the performance.

The transition matrix estimator highly depends on the posteriors from neural networks, when neural network changes the transition matrix changes accordingly. This can be a source of instability for our label noise learning. In this paper, we have used different estimated transition matrices for each iteration.

The performance of the neural network can be impacted by model tuning. To fairly compare the performance, we need to properly tune each of the models. For example using a smaller learning rate and assessing its convergence across epochs to select a final model without overfitting. However, due to time and computational limitations we are unable to do so for this paper.

\section{Conclusion and future Work}
In this paper, we have examined the fundamental concept underlying related label noise approaches. A transition matrix estimator has been built, and its efficacy versus the genuine transition matrix provided by the two FashionMINIST datasets has been demonstrated. In addition, we examined the label noise robustness of two convolutional neural network classifiers with LeNet and AlexNet designs. Regarding the two FashionMINIST datasets, both models have proven to be robust. We are not efficiently able to demonstrate the influence of the transition matrix noise correction on robustness improvements, however, due to our inability to correctly tune the complex convolutional neural network model due to time and computing resource constraints. To correctly tune the neural network model and analyse the accuracy of the estimated transition model, additional work is required.

In addition, we would like to make a few modifications to the experiment's design. The calculated transition matrix for CIFAR may contain errors in comparison to the actual matrix. If the transition matrix is not precisely predicted, the model's resilience could be severely diminished. To estimate the transition matrix in future work, additional refinement needs be conducted. In addition, we would like to investigate more transition matrix estimating techniques, such as T-revision \cite{xia_are_2019}.

\bibliographystyle{splncs04}
\bibliography{main_content}

\end{document}